# Time Series Classification to Improve Poultry Welfare

Alireza Abdoli [1], Amy C. Murillo [2], Chin-Chia M. Yeh [1], Alec C. Gerry [2], Eamonn J. Keogh [1]

[1] Department of Computer Science and Engineering, [2] Department of Entomology
University of California, Riverside, CA 92521
{aabdo002, amy.murillo, myeh003, alecg}@ucr.edu, eamonn@cs.ucr.edu

*Abstract* — **Poultry farms are an important contributor to the human food chain. Worldwide, humankind keeps an enormous number of domesticated birds (e.g. chickens) for their eggs and their meat, providing rich sources of low-fat protein. However, around the world, there have been growing concerns about the quality of life for the livestock in poultry farms; and increasingly vocal demands for improved standards of animal welfare. Recent advances in sensing technologies and machine learning allow the possibility of automatically assessing the health of *some* individual birds, and employing the lessons learned to improve the welfare for *all* birds. This task superficially appears to be easy, given the dramatic progress in recent years in classifying *human* behaviors, and given that human behaviors are presumably more complex. However, as we shall demonstrate, classifying chicken behaviors poses several unique challenges, chief among which is creating a generalizable "dictionary" of behaviors from sparse and noisy data. In this work we introduce a novel time series dictionary learning algorithm that can robustly learn from weakly labeled data sources.**

## I. Introduction

Domesticated birds, e.g. chickens, are a major source of food for humans; people have been keeping birds for their eggs (laying chickens) and for their meat (broiler chickens) for thousands of years [1]. Poultry farms are a major source of high-protein and low-fat food.

Given the ever-increasing population of the world, the demand for such food sources has been steadily growing. According to Food and Agriculture Organization of the United Nations (FAO), poultry meat consumption around the world has climbed from 11 kg/person in 2000 to 14.1 kg/person in 2011 [2]; and it is predicted to continue for the foreseeable future [3]. In developed countries, there are growing concerns about the ethical treatment of these animals; among which are housing conditions and how the animals are managed and treated [4].

Ectoparasites are a group of arthropods that reside on the surface of the body of chickens, causing stress to the host, and potentially spreading to nearby chickens or other animal hosts [37][39]. Many of these ectoparasites, such as the northern fowl mite, adversely affect productivity (e.g. laying eggs) and health of the chickens [14]. They may also impact poultry behavior and welfare [4][19]. Understanding how ectoparasites affect chicken behavior can help producers determine when flocks are infested, to better deploy control methods [15][39]. Traditional behavior studies have relied on direct or video observation of subjects. However, this is time consuming, error-prone and subjective.

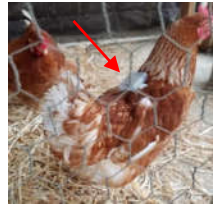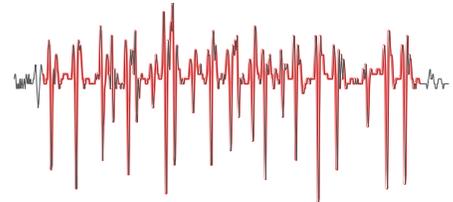

Seven seconds of X-axis chicken acceleration data

**Figure 1:** (*left*) A chicken with an Axivity AX3 accelerometer worn inside a 'backpack' on the back of chicken (*right*) A seven-second snippet of chicken time series data collected from the accelerometer.

We argue that the use of on-animal sensors can help to increase richness and density of observations [20]. Additionally, sensors can be used to greatly increase the number of individuals that can be tracked, while also expanding the tracking period, in some cases to 24/7 monitoring. In recent years, there have been enormous technological advances in sensor technology, and consequently sensor prices have decreased dramatically; this in turn has made sensor-driven data collection a practical option. Recently, there have been various studies on using sensors for collecting data in the context of livestock and poultry [5][6][18][20], even flying insects have not escaped the efforts at digitally sensed monitoring [7].

Such collected sensor data is typically extracted in the form of time series data; which can be examined offline with data mining techniques to summarize the behaviors of the animals under study. Time series data is widely used in many domains and is of significant interest in data mining [22]-[26]. In recent years researchers have proposed various algorithms for the efficient processing of time series datasets [10][17][27][28][32], even in case of noisy data [38]. In this study, we use on-animal sensors to quantify specific behaviors performed by chickens. These behaviors, e.g. *dustbathing*, are known or suspected to correlate with animal well-being [4].

Before moving on, we will take the time to ward off a possible misunderstanding. We are *not* proposing that all chickens be monitored, that is clearly unfeasible. Our system is designed as a tool to allow researchers to assess the effects of various conditions on chicken health, and then use the lessons learned on the entire brood. Our particular motivating example initiated with a study at UC Riverside to assess the effects of ectoparasites on chickens. However, our methods can be used to determine the effects of any change in the bird's environment or diet. Our only assumption is that the change will manifest in changes in the frequency or timing of the bird's behavior(s).

### A. Why is studying chicken behaviors at scale hard?

There are hundreds of studies on quantifying *human* body behaviors with sensors. Such studies typically involve finding discrete well-defined classes of behaviors, and then monitoring data for future occurrences of behaviors. One example is "step-counting" to measure compliance with a suggested exercise routine. However, the task of studying chicken behaviors is more difficult for the following reasons:

- In case of humans, the sensors can be easily placed on the extremities of the limbs (i.e. smart-shoes or smartwatches); However, the placement of sensors on chickens has been primarily restricted to the back of the birds, see Figure 1 (*left*), due to sensor limitations and the welfare of animal. This provides only coarse information about the bird's behaviors.

- The variability of human behaviors is well-studied, and it is understood what fraction can be attributed to individual personality, mood and so forth, versus variability in sensor placement [30]. More importantly, it is known how to account for this variability. However, it is less clear how much variability exists in birds and how we can best account for it.

- Creating a dictionary for a human subject is relatively straight-forward. During an explicit training session, behaviors of interest can be *acted out* in a fixed order for a fixed duration of time. Perhaps the most studied human motion time series dataset is *gun-point* [11]. When recording that dataset, the actor's behaviors were cued by a metronome. Chickens are clearly not as cooperative, and many hours/days of video recordings must be analyzed to create a behavior dictionary. Moreover, it is difficult, even for an experienced avian ethologist, to define precisely where a behavior begins and ends, thus we must be able to work with "weakly labeled" data.

In this work, we introduce a novel dictionary learning algorithm which can take weakly labeled data in the format "*there are a few pecks somewhere in this time period,*" together with some mild constraints "*a preening behavior probably lasts between 0.3 and 1.5 seconds*" and automatically construct a dictionary of behaviors. As we shall show, this dictionary can then be used to classify unlabeled archives of bird behavior.

The rest of this paper is organized as follows. In Section II we review related work and background material. Section III introduces our algorithm. An extensive empirical evaluation is conducted in Section IV. We provide a case study in Section V, before we offer conclusions and directions for future work in Section VI.

## II. RELATED WORK AND BACKGROUND

In recent works [12][18], sensors were used for classification of sheep behaviors; with mounted sensors on ear/collar or leg of the sheep. Similar work has been performed for domestic bovines [5], and for various kinds of wild animals. In addition, there has been some work on poultry behaviors using both sensors and human monitoring [34][35]. However, this work is complementary to our efforts. They use *statistical* features of the accelerometer (mean, entropy, etc.) to quantify periods of general behaviors, such as `sleep`, `stand`, `walk`, etc. [35]. In contrast, we use the *shape* of the time series to precisely annotate very specific and dynamic behaviors, such as individual instances of a single `peck`. By analogy with human studies, this is similar to recognizing the difference between when someone having lunch versus recognizing each individual bite. Both types of information can be useful for various tasks.

We begin by providing definitions and notation to be used throughout the paper.

### A. Definitions and Notation

**Definition 1:** A *time series* $T$ is a sequence of real-valued numbers $t_i$: $T = [t_1, t_2, …, t_n]$ where $n$ is the length of $T$.

We are typically not interested in the *global* properties of time series, but in the similarity between local *subsequences*:

**Definition 2:** A *subsequence* $T_{i,m}$ of a time series $T$ is a continuous subset of the values from $T$ of length $m$ starting at position $i$. $T_{i,m} = [t_i, t_{i+1}, …, t_{i+m-1}]$ where $1 \leq i \leq n - m + 1$

We can take any subsequences and calculate its distance to *all* subsequences in a time series. An ordered vector of such distances is called a *distance profile*:

**Definition 3:** A *distance profile D* is a vector of Euclidean distances between a given query and each subsequence in the time series.

Figure 2 illustrates calculating distance profile (*D*). It is assumed that the distance is measured using the Euclidean distance between Z-normalized subsequences [11][27][28]. In Figure 2, the query $Q$ is extracted from time series $T$. As can be seen, distance profile has low values at the location of subsequences which are highly similar to the query $Q$. In case the query $Q$ is taken from the time series itself, then the value for the distance profile at the location of query should be zero, and close to zero just before and just after. To avoid *such trivial matches* an exclusion zone with the length of ($m/2$) is placed to the left and right of the query location [27][28].

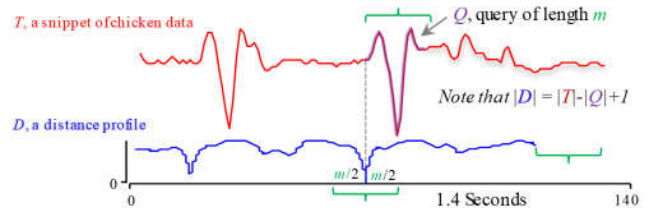

**Figure 2: Distance profile *D* obtained by searching for query *Q* (here the query is extracted from the time series itself) in time series *T*.**

The distance profile can be computed very efficiently using the MASS algorithm [8].

## III. DICTIONARY OF CHICKEN BEHAVIORS

### A. Collecting and Preparing Chicken Data

All chickens were housed and cared for in accordance with UC Riverside Institutional Animal Care and Use Protocol A-20150009. Data is collected from chickens by placing the sensor on bird's back, as shown in Figure 1 (*left*). The sensor is placed on back of the bird to allow for high-quality recording of various types of typical chicken behaviors, with the minimum interference and discomfort. The Axivity AX3 sensor used in our study, weighs about 11 grams and is configured with 100 Hz sampling frequency and +/- 8g sensitivity which allows for two weeks of continuous data collection with the battery fully charged.

From literature reviews [19][35][36], and conversations with poultry experts, we expect that the following behaviors correlate with poultry health:

- **Feeding/pecking:** bringing the beak to the ground; striking at the ground.
- **Preening:** preening/grooming of the feathers by the beak; feathers may be drawn or nibbled by the beak [21].
- **Dustbathing:** bird is in a sitting or lying position with feathers raised in a vertical wing-shake [21].

Figure 3 (*top left*) shows the Axivity AX3 sensor secured inside a plastic backpack with a rubber band wrist to allow placing the backpacks on the back of chickens (*right*) shows placement of sensor on the back of the chicken.

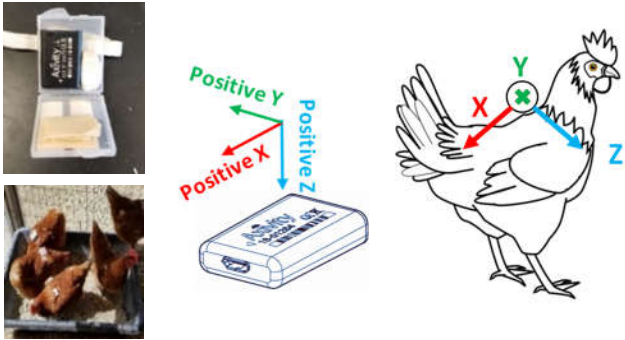

**Figure 3:** (*top left*) Axivity AX3 sensor secured inside a plastic backpack with a rubber band wrist (*bottom left*) Chickens wearing backpacks on their back (*center*) Axivity AX3 axis alignment (*right*) positioning of AX3 sensor on the back of chicken.

For some fraction of the time series data collected, a video camera also recorded the chicken activity (~ 30 minutes). This provides ground truth to act as training data. The sensor data was carefully annotated [13], based on the video-recorded chicken activities by team member A.C. M; it must be noted that even the most careful human labeling of chicken behaviors can contain errors, especially false negatives.

More importantly, due to technical limitations it is difficult to synchronize the two data sources to anything less than one-second variable lag, which is a long time relative to a chicken peck (~ ¼ of a second). Thus, as shown in Figure 4, the annotations take the form of a categorical vector (shown in green) that indicate that in the corresponding region one or more examples of the corresponding behavior were observed. Such data is often called "weakly labeled" data.

To reiterate, inspecting this video is very time consuming for a technician; we do not propose to inspect *all* data this way, this is simply a one-time ground truthing operation performed for a tiny fraction of the data collected.

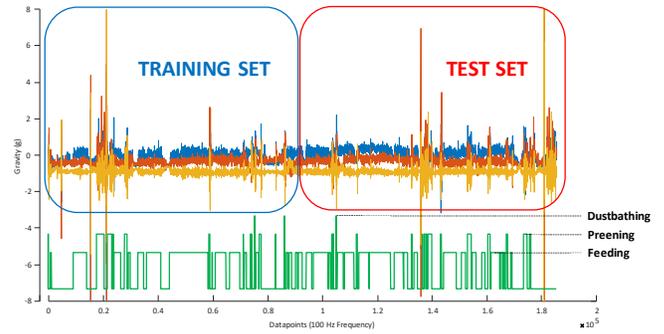

**Figure 4:** Three-dimensional chicken time series (the top/blue time series is X-axis; middle/red time series is Y-axis and bottom/orange time series is Z-axis time series). The green lines represent annotations of observed chicken behaviors captured on video; the height of each annotated region represents a distinct behavior.

*B. Creating a Dictionary of Behaviors*

Given that we have training/test data in the format shown in Figure 4, we are in a position to attempt to automatically construct a *dictionary* of behaviors or query-templates.

A dictionary is a list of query-templates (behaviors) in the form of $\{s_1, s_2, …, s_i\}$; each query has a class ($s_1.class$), a threshold value ($s_1.threshold$) and axis ($s_1.axis$) properties, along with the query data points.

In principle, a single behavior could have two or more possible instantiations; just like the number four has two written versions, closed '4' and open '4', which are semantically identical. Such a dictionary is called a polymorphic dictionary. Given our observations of chicken behaviors, in this work we assume that there is a single way to perform a behavior. However, generalizing the code to a polymorphic dictionary is trivial and omitted for brevity here.

*C. Algorithms for Building the Chicken Behavior Dictionary*

Our algorithm for building a dictionary of chicken behaviors works by searching within annotated regions for highly conserved sequences (i.e. *motifs* [27][28]). To give an intuition for this, consider the analogue problem in the discrete string space. Imagine we are given this data snippet, and we are told that the green region is a weak label for *pecking*.

`dbehiorhfbesoqhebesoweqhfebesopwehfuwbeibe`

Further suppose we are told that the length of the behavior is between 2 and 5 symbols. If we looked for conserved behavior of length 2, we would find that 'be' happens six times. However, three of those occurrences are outside the annotated region, so this cannot be a good predictor of the class. If we looked for conserved behavior of length 3, we would find that 'bes' happens three times, and all of the occurrences happen within the labeled region. This seems like a better predictor of the class. However, note that if we expand to conserved pattern of length 4, 'beso' also happens three times within the labeled region. Since we might expect that the longer pattern is more specific to the class, we prefer it. Note that if we continue the search to patterns of length five, there are no highly conserved patterns of this length.

In addition to time series *T* and the annotation labels *Label*, the algorithm takes a range of lengths *Len* (equivalent to the values 2 to 5 in the example above). TABLE I. presents the pseudo-code for building a dictionary of behaviors.

In Line 1 the range of query-template lengths to be tested is specified. Line 2 iterates over the annotated regions. In Line 3, a sliding window is initiated inside the selected region from Line 2 with the specified length from Line 1. The selected query-template, time series, label data are provided to the nearest neighbor similarity search subroutine algorithm (TABLE II. ). The output result for the selected query-template is added to the current list of query-templates (Q*ueryList*) in Line 6.

**TABLE I.**   ALGORITHM FOR BUILDING DICTIONARY

| Procedure DictionaryBuilder (*T, Label, Len*) |
|---|
| **Input:** Time Series (*T*), researcher labels (*Label*) and length range (*Len*) |
| **Output:** List of query-templates (*QueryList*) |
| 1   **for** *QLen = Len* |
| 2    **for** *LblRegion* = 1:length(*Label*) |
| 3     **for** *Q* = Label (*LblRegion*, 1): Label (*LblRegion*, 2) |
| 4      [*K, dist, TP, FP*] = |
| 5         NN (*T$_{Train}$, T$_{Train}$(Q:Q + QLen - 1), Label*) |
| 6      *QueryList* = [*QueryList*; *Q, K, dist, TP, FP*] |
| 7     **end** |
| 8    **end** |
| 9   **end** |
| 10  **return** *QueryList* |

TABLE II. is a subroutine, which given the candidate query-template, time series and label data calculates the similarity between the query-template and all subsequence in the time series.

**TABLE II.**   NEAREST NEIGHBOR SIMILARITY SEARCH ALGORITHM

| Procedure NN (*T, Q, Label*) |
|---|
| **Input:** Time Series (*T*), a query (*Q*), and entomologist labels (*Label*) |
| **Output:** No. of True Positives (*TP*), No. of False Positives (*FP*) and Distance value (*dist*) |
| 1   *D* ← MASS (*T, Q*) // see [8] |
| 2   [*D$_s$, D$_{idx}$*] ← sort (*D*) // sorts ascendingly and returns indices |
| 3   *Idx* = 1,  *TP* = 0,  *FP* = 0, *dist* = 0; |
| 4   **if** (Label (*D$_{idx}$(Idx)*) == True) |
| 5    *TP* = *TP* + 1 |
| 6    *dist* = *D$_s$(Idx)* |
| 7   **else** |
| 8    *FP* = *FP* + 1 |
| 9   **end** |
| 10  **if** (FP == 0) |
| 11    *Idx* = *Idx* + 1 |
| 12  **else** |
| 13    **return** |
| 14  **end** |

In Line 1, the candidate query-template and time series data are passed to the MASS algorithm. MASS computes the similarity between query-template (*Q*) and every subsequence in time series (*T*); and returns a distance profile *D* [8]. The distance profile *D* is sorted in ascending order (*D$_s$*) and the indices of sorted values are stored in *D$_{idx}$* in Line 2.

In Lines 4 - 9, at each iteration, the algorithm takes a value from (*D$_{idx}$(Idx)*) and examines the corresponding value of *Label*(*D$_{idx}$(Idx)*) to see if it is marked as a targeted behavior. If it is marked as a behavior of interest, then the subsequence is a true positive (*TP*) match and *TP* is increased by one (Line 5), otherwise, it is treated as a false positive (*FP*) match and *FP* is increased by one (Line 8). We are interested in candidate query-templates which yield the highest number of true positive matches with no false positive matches. Therefore, in Lines 10 - 14, we continue the search as long as no false positive (*FP*) match is observed. It is worth noting that the distance (*dist*) mentioned in Line 6 of TABLE II is the same as query-template *threshold* value in the behavior dictionary. The *threshold* value serves as a measure of similarity when searching in unlabeled data streams. In the case, a subsequence in the unlabeled data has a similarity value with the query-template lower than the threshold it is classified as a matching subsequence.

*D. How to use the Chicken Behavior Dictionary?*

When monitoring the stream of time series data, we look for specific behaviors to count, classify, and time-stamp. The stream of data is processed through a sliding window. In case the similarity *threshold* is met for some query-template (behavior) in the dictionary, then the sequence is matched and time-stamped as an instance of that behavior.

IV. EMPIRICAL EVALUATION

To ensure that our experiments are reproducible, we have built a supporting website [31]; which contains all data, code and raw spreadsheets for the results, in addition to many experiments that are omitted here for brevity.

We provide evaluation results for the `feeding/pecking`, `preening`, and `dustbathing` behaviors. The original dataset is split into mutually exclusive training and test datasets, as illustrated in Figure 4. The training dataset is used for building the dictionary of behaviors, while the test dataset is solely used for out-of-sample evaluation. As Figure 4 reminds us, the data is weakly labeled, meaning that every annotated region contains one or more of the specified behavior. In addition, there are almost certainly instances of the behavior outside the annotated regions which the annotator failed to label, perhaps because the chicken in question was occluded in the video. However, we believe that such false negatives are rare enough to be ignored.

In addition, we do *not* know about the exact number of individual instances of a behavior inside a region, complicating the evaluation. To address this, we utilize the concept of Multiple Instance Learning (MIL) [29]; which assumes each annotated region as a "bag" containing one or more instances of a behavior. In this classification model, if at least a single instance of a behavior is matched inside a bag, it is treated as a true positive. However, in case that no instances of the behavior are detected inside the bag, then the entire bag is treated as a false negative. Furthermore, in case an instance of behavior is mismatched inside a bag corresponding to some other behavior, then it is treated as a false positive. Finally, if no mismatch occurs inside a bag of non-relevant behavior, then the entire bag is treated as a true negative.

## A. Feeding/Pecking Behavior

Feeding/pecking is perhaps the most familiar behavior in chickens. Figure 5 shows the query-template discovered by our dictionary building algorithm, along with matching subsequences in the *training* dataset. Recall that subsequences located within regions annotated as containing instances of feeding/pecking behavior are treated as true positives (TP), whereas matches outside of regions annotated for feeding/pecking behavior are treated as false positives (FP).

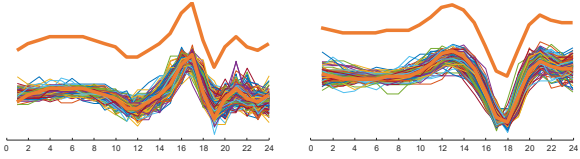

**Figure 5: Query-template for feeding/pecking behavior, (*left*) X-axis and (*right*) Z-axis matching subsequences in the training dataset.**

Figure 6 shows matching subsequences from the *test* dataset with true positives shown in green and false positives shown in red. The reader will appreciate that the false positives *do* look at lot like the true positives. As noted above, it is possible (and indeed *likely*) that they really are true positives that escaped the attention of the human annotator.

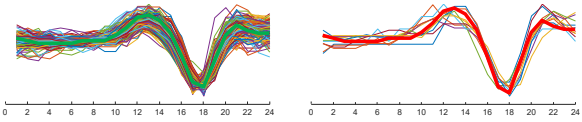

**Figure 6: The X-axis query-template and matching subsequences for feeding/pecking behavior in the test dataset, (*left*) true positives (*right*) false positives. Note that the false positives do *appear* to actually be true positives that were missed by the human annotator.**

Given the positioning of sensor on the back of chicken, as shown in Figure 3 (*center*) and (*right*), when the chicken approaches its head to the ground, X-axis and Z-axis are affected. As can be seen in Figure 5 (*top*), pecking behavior is manifested in the form of a "valley" on X-axis (i.e. negative X-axis acceleration); at the same time as shown in Figure 5 (*bottom*), the pecking behavior is manifested in the form of a peak on Z-axis (i.e. positive Z-axis acceleration). Further, note that Y-axis (i.e. lateral acceleration towards left or right) is not as influential as X and Z-axis, therefore we omit the Y-axis for this specific behavior. Figure 8 (*top*) shows the matching subsequences for running the feeding/pecking query-template against the test dataset. TABLE III. provides the confusion matrix for the performance of feeding/pecking query-template.

**TABLE III.** CONFUSION MATRIX FOR FEEDING/PECKING BEHAVIOR

|  |  | Actual Class | |
|---|---|---|---|
|  |  | **Feeding** | **Non-Feeding** |
| **Predicted Class** | **Feeding** | 17 True Positives | 7 False Positives |
|  | **Non-Feeding** | 4 False Negatives | 43 True Negatives |

$$\text{Precision}_{(Feeding/Pecking)} = \frac{TP}{TP + FP} = 17 / 24 = 0.71$$

$$\text{Recall}_{(Feeding/Pecking)} = \frac{TP}{TP + FN} = 17 / 21 = 0.81$$

$$\text{Accuracy}_{(Feeding/Pecking)} = \frac{TP + TN}{TP + FP + TN + FN} = 60 / 71 = 0.85$$

Given the results above, our classification model has a 0.71 precision and 0.81 recall in matching instances of the feeding/pecking behavior. Overall, the classifier has 85% accuracy for the feeding/pecking behavior, which compares very favorably to 70% default rate (i.e., guessing every observed object as the majority class).

## B. Preening Behavior

Preening is the act of cleaning feathers, which is an important part of a chicken's daily activities. Preening is a grooming behavior that involves the use of the beak to position feathers, interlock feather barbules (informally, "zip up" the feathers) that have become separated, clean plumage, and remove ectoparasites [16]. Figure 7 shows the query-template for the preening behavior and the matching subsequences within the training dataset. Also, Figure 8 shows subsequences for running the query-template against a long test dataset.

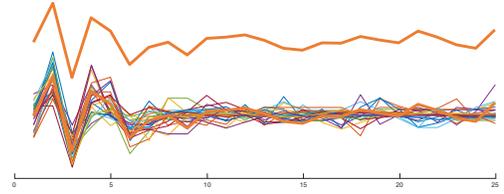

**Figure 7: The Z-axis query-template corresponding to preening behavior along with matching subsequences in the training dataset.**

It is interesting to discuss the nature of the preening behavior before proceeding to evaluation results. One might imagine that a flat region in the data is uninformative, however a flat region directly after the pattern on the Figure 7 *is* informative. The bird uses its beak to realign the feathers, and as can be seen in Figure 7, the act of preening starts with moving the head (i.e. positive and negative Z-axis acceleration) and the rest of the pattern is relatively flat which might imply movement of feather through the beak for alignment purposes. Figure 8 (*bottom*) shows the matching subsequences for running the preening query-template against the test dataset. TABLE IV. provides the confusion matrix on the performance of preening query-template.

**TABLE IV.** CONFUSION MATRIX FOR PREENING BEHAVIOR

|  |  | Actual Class | |
|---|---|---|---|
|  |  | **Preening** | **Non-Preening** |
| **Predicted Class** | **Preening** | 10 True Positives | 1 False Positives |
|  | **Non-Preening** | 4 False Negatives | 56 True Negatives |

$$\text{Precision}_{(Preening)} = 10 / 11 = 0.91$$
$$\text{Recall}_{(Preening)} = 10 / 14 = 0.71$$
$$\text{Accuracy}_{(Preening)} = 66 / 71 = 0.93$$

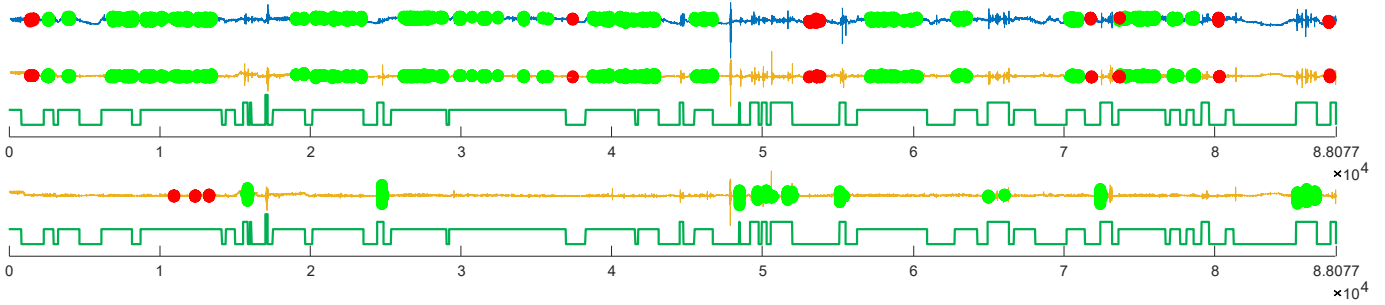

**Figure 8:** Matching subsequences in the test dataset for (*top*) feeding/pecking behavior on X and Z-axis (*bottom*) preening behavior only on Z-axis. **KEY: green** highlightings indicate true positives, and **red** highlightings indicate false positives.

Given the evaluation results above our classification model has 0.91 precision in matching preening subsequences; and 0.71 recall in matching relevant instances of the preening behavior. Finally, the model has 93% overall accuracy in matching preening subsequences compared to the 80% default rate (i.e., guessing every observed object as the majority class).

### C. Dustbathing Behavior

Dustbathing is the act in which the chicken moves around in dust or sand to remove parasites from its feathers. This tends to be the least common activity in chickens. As shown in Figure 9 and Figure 11, dustbathing tends to be the most difficult behavior to search for, since there were only two instances in the training dataset and only a single instance in the test dataset.

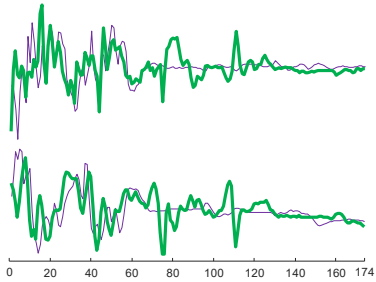

**Figure 9:** Matching subsequence for dustbathing behavior in the test dataset, (*top*) X-axis (*bottom*) Z-axis.

TABLE V. presents the confusion matrix for the dustbathing behavior.

**TABLE V.** CONFUSION MATRIX FOR PREENING BEHAVIOR

|  |  | Actual Class | |
|---|---|---|---|
|  |  | **Dustbathing** | **Non-Dustbathing** |
| **Predicted Class** | **Dustbathing** | 1 True Positives | 0 False Positives |
|  | **Non-Dustbathing** | 0 False Negatives | 70 True Negatives |

$$\text{Precision }_{(Dustbathing)} = 1 / 1 = 1.00$$
$$\text{Recall }_{(Dustbathing)} = 1 / 1 = 1.00$$
$$\text{Accuracy }_{(Dustbathing)} = 71 / 71 = 1.00$$

Given these evaluation results, our model has 1.00 precision in matching dustbathing subsequences and 1.00 recall in matching relevant instances of the dustbathing behavior. Finally, the model has 100% overall accuracy in matching dustbathing subsequences compared to 99% default rate (i.e., guessing every observed object as the majority class). Figure 11 shows subsequences for running the dustbathing query-template against the test dataset.

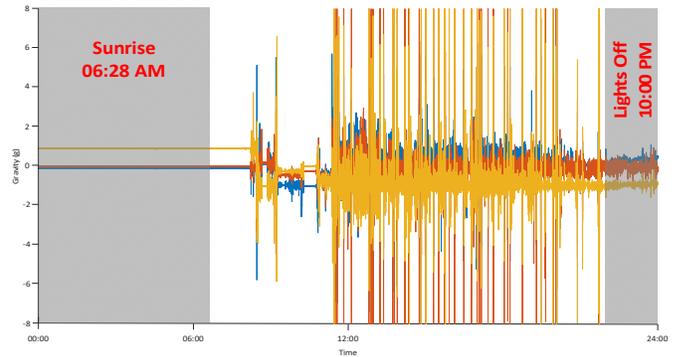

**Figure 10:** Twenty-four hours time series chicken data.

### V. CASE STUDY: AN ENTIRE DAY WITH A CHICKEN

In this section, we study the behavior of a chicken over the course of 24 hours and run our chicken behavior dictionary against the day-long dataset. The data corresponds to 11/22/2017 from midnight to midnight. The dataset is shown in Figure 10; and is of size 8,665,227 x 3 datapoints. The gray shaded regions correspond to midnight to sunrise (i.e. 06:28 AM) and the time artificial lights are turned off (i.e. 10:00 PM) to midnight.

We ran the dictionary of chicken behaviors against the twenty-four hours chicken dataset. Figure 12, Figure 13, and Figure 14 present the matching subsequences for feeding/pecking, preening, and dustbathing query-template behaviors in the 24 hours chicken dataset.

Starting with Figure 12, we show the matching subsequences in both the original and Z-normalized space to justify our choice of working with the Z-normalized representation [8][10]. Note that in the original space there are shifts in the mean that are inconsequential, yet which would dwarf the Euclidian distance calculations [9][11].

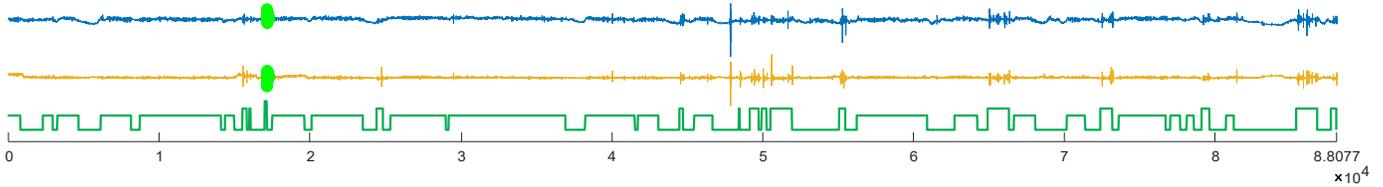

Figure 11: Matching subsequence in the test dataset for dustbathing behavior, while this behavior is much rare, our algorithm correctly found the instance. KEY: green highlightings indicate true positives, and red highlightings indicate false positives (we do *not* have any false positive matches here).

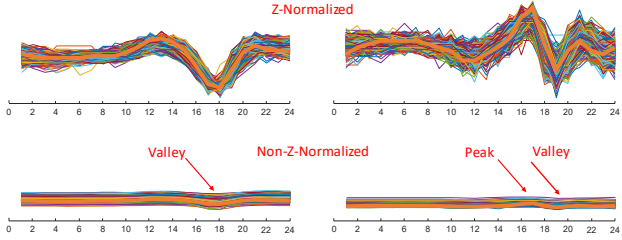

Figure 12: The X-axis and Z-axis query-templates, along with matching subsequences for feeding/pecking behavior in the twenty-four hours chicken dataset, (*top*) Z-Normalized and (*bottom*) Non-Z-Normalized matching subsequences.

Looking at Figure 13 for the query-template and matching subsequences for the preening behavior, it can be seen that the matching subsequences have very high similarity with the query template at the beginning; however, we still see some dissimilarities along the rest of the query-template which may correspond to nibbling of the feathers with the beak.

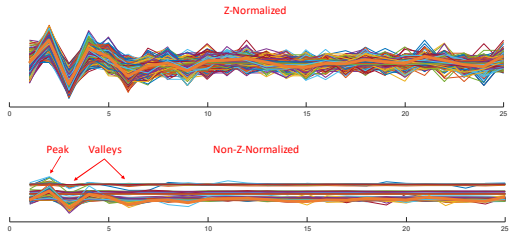

Figure 13: The Z-axis query-template and matching subsequences for the preening behavior in the twenty-four hours chicken dataset, (*top*) Z-Normalized matching subsequences and (*bottom*) Non-Z-Normalized.

Finally in Figure 14 the matching subsequences for the dustbathing behavior are shown. The matching subsequences are fairly well conserved, however, not as well as the feeding/pecking and preening behaviors. This can be due to the fact that dustbathing is a relatively long (~1.74 s) and a rare behavior.

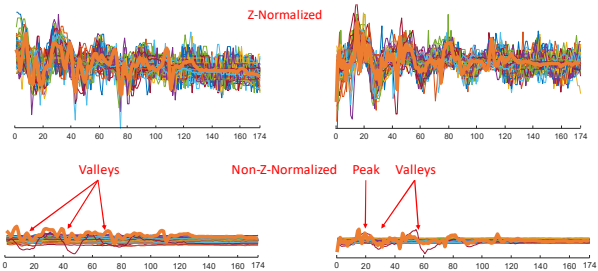

Figure 14: The X-axis and Z-axis query-templates, along with matching subsequences for dustbathing behavior in the twenty-four hours chicken dataset (*top*) Z-Normalized (*bottom*) Non-Z-Normalized matching subsequences.

Figure 15 shows the frequency of each behavior over the entire 24 hours, as computed with a one-hour long sliding window. As expected, the feeding/pecking behavior has the highest overall frequency; peaking between 11:00 AM and 15:00 PM. An interesting finding is that the chicken seems to begin the day with a preening session starting just after dawn. A more unexpected finding is that there is a "dip" in feeding that happens just after noon, and it seems to be replaced with an uptick in dustbathing. In future work we will examine similar circadian traces to understand if this is typical of all birds or indicative of an individual with an ectoparasite infestation.

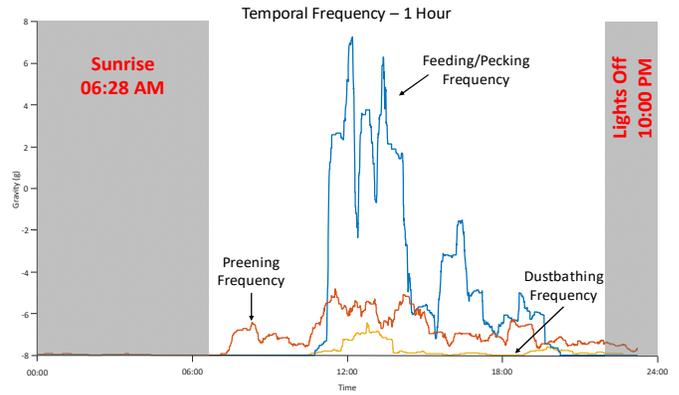

Figure 15: Temporal frequency of chicken behaviors (feeding/pecking, preening and dustbating) for the twenty-four hour chicken dataset.

## VI. CONCULSION

In this work, we introduced an algorithm to learn a dictionary of behaviors from weakly labeled time series data. We demonstrated, with an extensive empirical study, that our algorithm can robustly learn from real, noisy and complex datasets, and that the learned query-templates generalize to previously unseen data. While our study was motivated by a pressing problem in poultry welfare, it could clearly be used to study cattle [5][21], goats [34], non-human primates [33], or any other data source that presents itself as weakly labeled data. In future and ongoing work, we plan to examine much larger datasets; corresponding to several years of chicken data (recorded in parallel from multiple birds).

ACKNOWLEDGMENT

We would like to acknowledge funding from "NSF 1510741–RI: Medium: Machine Learning for Agricultural and Medical Entomology (07/01/17-09/30/19)" and "USDA NIFA-2017-67012-26100".